\documentclass[11pt,a4paper]{article}
\usepackage[hyperref]{emnlp2020}
\usepackage{times}
\usepackage{latexsym}
\usepackage{graphicx}

\usepackage{microtype}
\usepackage{amsmath,amssymb,amsfonts}
\usepackage[mathscr]{euscript}
\usepackage{amsbsy}
\usepackage{multirow}
\usepackage{booktabs}
\usepackage{subcaption}
\usepackage{commath}
\usepackage{newtxtext,newtxmath}
\usepackage{relsize}
\usepackage{dsfont}

\aclfinalcopy 

\DeclareMathAlphabet\mathbfcal{OMS}{cmsy}{b}{n}

\title{Structured Self-Attention Weights \\ Encode Semantics in Sentiment Analysis}

\author{Zhengxuan Wu$^{1}$, Thanh-Son Nguyen$^{2}$, Desmond C. Ong$^{2,3}$ \\
  $^{1}$Symbolic Systems Program, Stanford University \\
  $^{2}$Institute of High Performance Computing, Agency for Science, Technology and Research, Singapore \\
  $^{3}$Department of Information Systems and Analytics, National University of Singapore \\
  {\tt wuzhengx@stanford.edu, Nguyen\_Thanh\_Son@ihpc.a-star.edu.sg,}\\ {\tt dco@comp.nus.edu.sg} }

\date{}

\begin{document}
\maketitle
\begin{abstract}
Neural \textit{attention}, especially the self-attention made popular by the Transformer, has become the workhorse of state-of-the-art natural language processing (NLP) models. Very recent work suggests that the self-attention in the Transformer encodes syntactic information; Here, we show that self-attention scores encode semantics by considering sentiment analysis tasks. In contrast to gradient-based feature attribution methods, we propose a simple and effective Layer-wise Attention Tracing (LAT) method to analyze structured attention weights. We apply our method to Transformer models trained on two tasks that have surface dissimilarities, but share common semantics---sentiment analysis of movie reviews and time-series valence prediction in life story narratives. Across both tasks, words with high aggregated attention weights were rich in emotional semantics, as quantitatively validated by an emotion lexicon labeled by human annotators. Our results show that structured attention weights encode rich semantics in sentiment analysis, and match human interpretations of semantics.
\end{abstract}

%
%
%

\section{Introduction}

In recent years, variants of neural network attention mechanisms such as local attention~\cite{bahdanau2015neural, luong2015effective} and self-attention in the Transformer~\cite{vaswani2017attention} have become the \emph{de facto} go-to neural models for a variety of NLP tasks including machine translation~\cite{luong2015effective, vaswani2017attention}, syntactic parsing~\cite{vinyals2015grammar}, and language modeling~\cite{liu2018learning, dai2019transformer}. 



Attention has brought about increased performance gains, but what do these values `mean'? Previous studies have visualized and shown how learnt attention contributes to decisions in tasks like natural language inference and aspect-level sentiment~\cite{lin2017structured, wang2016attention, ghaeini2018interpreting}. Recent studies on the Transformer~\cite{vaswani2017attention} have demonstrated that attention-based representations encode syntactic information~\cite{tenney2019bert} such as anaphora~\cite{voita2018context, goldberg2019assessing}, Parts-of-Speech~\cite{vig2019analyzing} and dependencies~\cite{raganato2018analysis, hewitt2019structural, clark2019what}. Other researchers have also done very recent extensive analyses on self-attention, by, for example, implementing gradient-based Layer-wise Relevance Propagation (LRP) method on the Transformer~\cite{voita-etal-2019-analyzing} to study attributions of graident-scores to heads, or graph-based aggregation method to visualize attention flows~\cite{abnar-zuidema-2020-quantifying}. These very recent works have not looked at whether the structured attention weights themselves aggregate on tokens with strong semantic meaning in tasks such as sentiment analysis. Thus, it is still unclear if the attention on input words may actually encode semantic information relevant to the task.

In this paper, we were interested in extending previous studies on attention and syntax further, by probing the structured attention weights and studying whether these weights encode task-relevant semantic information. In contrast to gradient-based attribution methods~\cite{voita-etal-2019-analyzing}, we were explicitly interested in probing learnt attention weights rather than analyzing gradients. To do this, we propose a Layer-wise Attention Tracing (LAT) method to aggregate the structured attention weights learnt by self-attention layers onto input tokens. We show that these attention scores on input tokens correlate with an external measure of semantics across two tasks: a sentiment analysis task on a movie review dataset, and an emotion understanding task on a life stories narrative dataset. These tasks differ in structure (single-example classification vs. time-series regression), and in domain (movie reviews vs. daily life events), but should share the same semantics, in that the same words should be important in both tasks. We propose a method of external validation of the semantics of these tasks, using emotion lexicons. We find evidence for the hypothesis that if self-attention mechanisms can learn emotion semantics, then LAT-calculated attention scores should be higher for words that have stronger emotional semantic meaning.~\footnote{Code is available at \url{https://github.com/frankaging/LAT_for_Transformer}}



\section{Attention-based Model Architecture}

\begin{figure}
    \centering
    \includegraphics[width=0.48\textwidth]{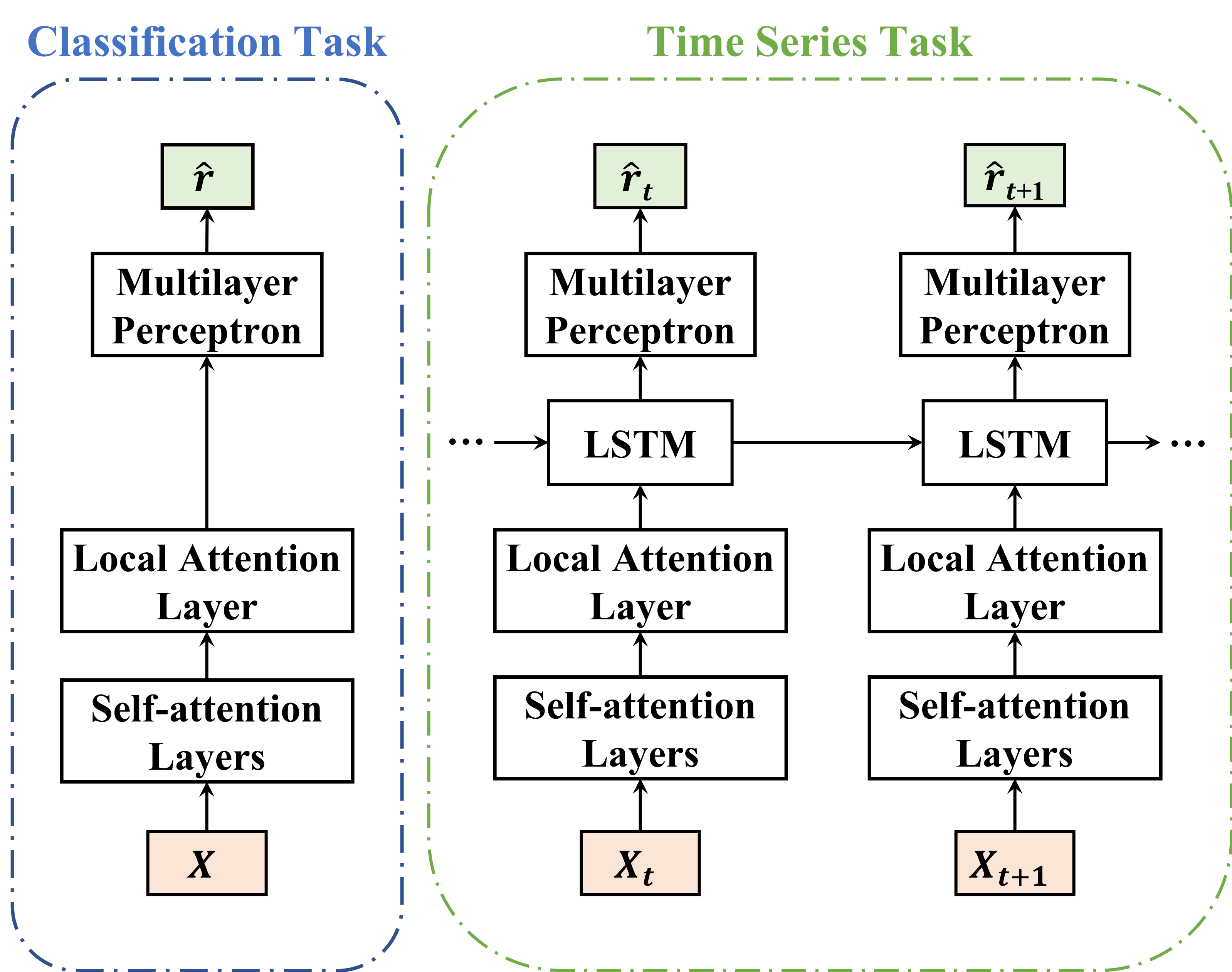}
    \caption{Attention-based encoder-decoder model architecture for classification task (left) and time-series task (right); The latter has a recurrent unit to generate predictions over time.}
    \label{fig:model}
\end{figure}

We use an encoder-decoder architecture as shown in Fig.~\ref{fig:model}. Our encoder is identical to the encoder of the Transformer~\cite{vaswani2017attention}, with an additional local attention layer~\cite{luong2015effective}. 
Our decoder is task-specific: a simple Multilayer Perceptron (MLP) for the classification task, and a LSTM followed by a MLP for the time-series prediction task. 

\paragraph{Self-attention Layers.}
%
%
The encoder is identical to the original Transformer encoder and consists of a series of stacked self-attention layers.
Each layer contains a multi-head self-attention layer, followed by an element-wise feed forward layer and residual connections. Following~\citet{vaswani2017attention}, we use $L = $ 6 stacked layers and $H = $ 8 heads, and a hidden dimension of $D = $ 512. 

We briefly recap the Transformer equations, to better illustrate our LAT method, which traces attention back through the layers. For a given self-attention layer $l \in [1,L]$, we denote the input to $l$ using $\textbf{\textit{X}}^{l} \in \mathbb{R}^{N \times D_e}$, which represents $N$ tokens, each embedded using a $D_e$-dimensional embedding. We keep the same input embedding size for all layers. The first layer takes as input the word tokens. 
A self-attention layer learns a set of \emph{Query}, \emph{Key} and \emph{Values} matrices that are indexed by $l$ (i.e., weights are not shared across layers). Formally, these matrices are produced in parallel:
\begin{align}
    \textbf{\textit{Q}}^{l} = f_{q}^{l} (\textbf{\textit{X}}^{l}) ,\enskip
    \textbf{\textit{K}}^{l} = f_{k}^{l} (\textbf{\textit{X}}^{l}) ,\enskip
    \textbf{\textit{V}}^{l} = f_{v}^{l} (\textbf{\textit{X}}^{l}) \label{eqn:first}
\end{align}
where $f_{\{q,k,v\}}^{l} (\cdot)$ are each parameterized by a linear layer, and each matrix is of size $N \times D$. To enable multi-head attention, $\textbf{\textit{Q}}$, \textbf{\textit{K}} and $\textbf{\textit{V}}$ are partitioned into $H$ separate $N \times D_{h}$ attention heads indexed by $h \in [1, H]$, where $D_{h} = \frac{D}{H} = $ 64.

Each head learns a self-attention matrix $\boldsymbol\alpha_{h}^{\textit{s}(l)}$ using the scaled inner product of $\textbf{\textit{Q}}_h$ and $\textbf{\textit{K}}_h$  followed by a \emph{softmax} operation. The self-attention matrix $\boldsymbol\alpha^{\textit{s}(l)}_{h}$ is then multiplied by  $\textbf{\textit{V}}_h$ to produce  $\textbf{\textit{Z}}_{h}^{l}$:
\begin{align}
    \boldsymbol\alpha^{\textit{s}(l)}_{h} &= \text{softmax}\left(\frac{\textbf{\textit{Q}}_{h}^{l}{\textbf{\textit{K}}^{l}_{h}}^{T}}{\sqrt{D_{h}}} \right) \label{eqn:linear} \quad &\in \mathbb{R}^{N \times N} \\ 
    \textbf{\textit{Z}}_{h}^{l} &= \boldsymbol\alpha^{\textit{s}(l)}_{h} \textbf{\textit{V}}_{h}^{l} \quad &\in \mathbb{R}^{N \times D_{h}} \label{eqn:self-v} 
\end{align}
Next, we concatenate $\textbf{\textit{Z}}_{h}^{l}$ from each head $h$ to produce the output of layer $l$ (i.e., the input to layer $l+1$) $\textbf{\textit{X}}^{l+1}$,
\begin{align}
    \textbf{\textit{X}}^{l+1} &= f_{\psi}^l ([\textbf{\textit{Z}}_{1}^{l}, ..., \textbf{\textit{Z}}_{H}^{l}]) \quad &\in \mathbb{R}^{N \times D_{e}} \label{eqn:ffn} 
\end{align}
where $f_{\psi}^l (\cdot)$ is parameterized by two fully connected feed-forward layers (with 64 dimensions for the first layer then scaling back to $D_{e}$-dimensions) with residual connections and layer normalization. $\textbf{\textit{X}}^{l+1}$ is fed upwards to the next layer.


\paragraph{Local Attention Layer.} The output from the last self-attention layer $\textbf{\textit{X}}^{L+1}$ is fed into a local attention layer. We then take a weighted sum over row vectors of the output, and produces a context vector $c$ using learnt local attention vector $\alpha^{\textit{c}}$:
\begin{align}
    \alpha^{\textit{c}} &= \text{softmax} \left( f_{\phi}(\textbf{\textit{X}}^{L+1}) \right) \label{eqn:local-attn} &\in \mathbb{R}^{N}\\ 
    c &= ( \textbf{\textit{X}}^{L+1} )^T \alpha^{\textit{c}} = \sum_{i=1}^{N} \alpha^{\textit{c}}_{i} {x_{i}^{L+1}} &\in \mathbb{R}^{D_{e}} \label{eqn:ctx-c}
\end{align}
where $f_{\phi}(\cdot)$ is parameterized by a multi-layer perceptron (MLP) with two hidden layers that are 128-dimensional and 64-dimensional. The MLP layers are trained with dropout of $p = $ 0.3.

\paragraph{Decoder.} For the classification task, the context vector $c$ is fed into a decoder $f_{dec}(\cdot)$ parameterized by a MLP to produce the output label. For the time-series task, context vectors $c_{t}$ from each time $t$ are fed into a LSTM~\cite{hochreiter1997long} layer with 300-dimensional hidden states before passing through a MLP. Both the MLP for the classification and the time-series tasks have the same 64-dimensional hidden space, and are trained with dropout of $p = $ 0.3. A complete model description can be found in the Appendix.

 \section{Layer-wise Attention Tracing}\label{sec:lat}
\begin{figure}[tb]
    \centering
    \includegraphics[width=\columnwidth]{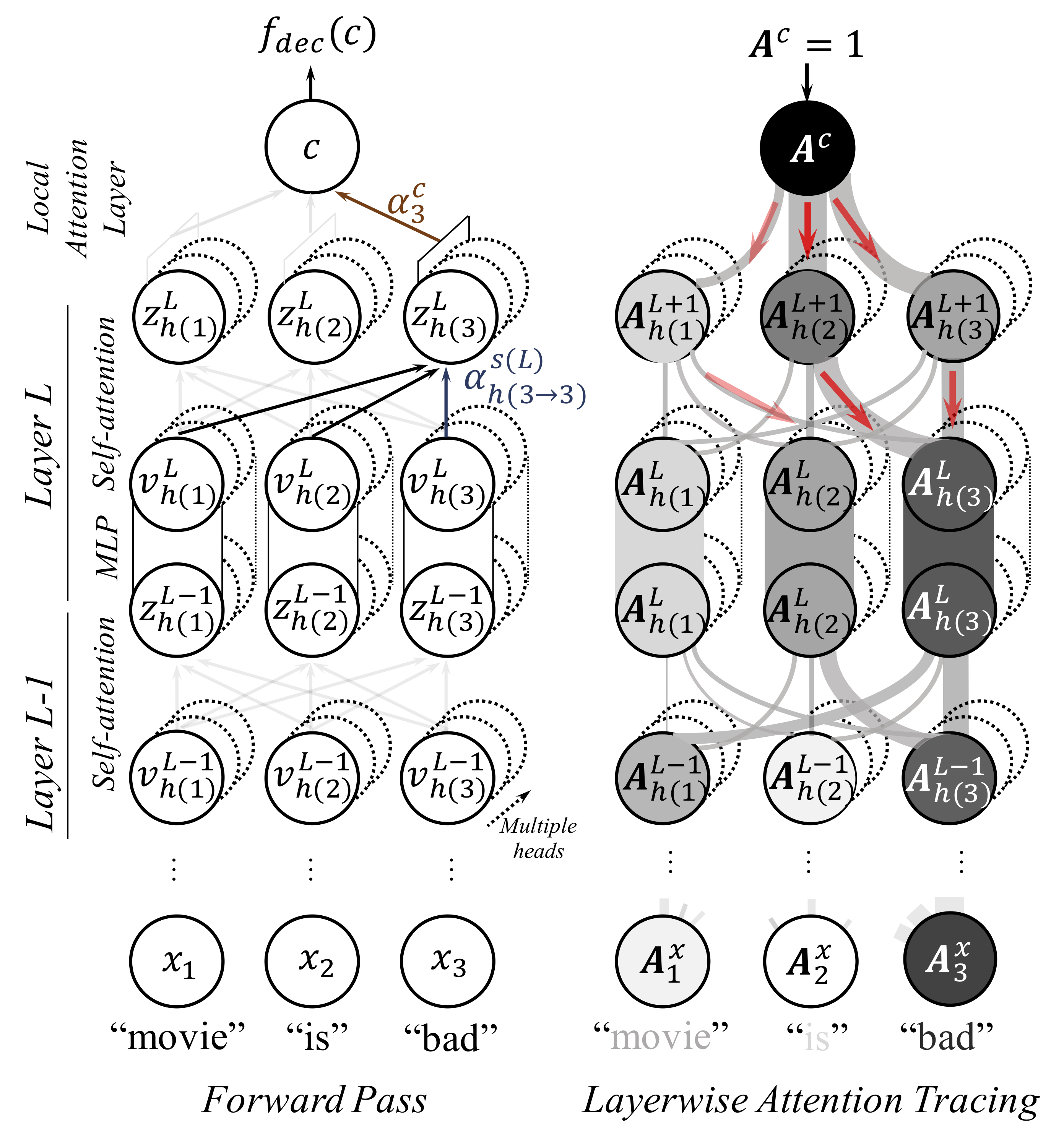}
    \caption{An illustration of the Layer-wise Attention Tracing (LAT) method with an example forward pass through head $h$. Left: On the forward pass, learnt attention weights are represented by lines producing $\textbf{\textit{Z}}_{h}^{l}$ from values $\textbf{\textit{V}}_{h}^{l}$ via self-attention (Eqn.~\ref{eqn:linear}-\ref{eqn:self-v}) and the context vector $c$ from the last layer via local attention (Eqn.~\ref{eqn:local-attn}). Dashed circles represents multiple heads, and vertical columns represent MLP transformations, which do not redistribution attention. Right: LAT on a `backward pass'. The thickness of the edges represents accumulating attention. Attention from incoming edges are accumulated at each position in each layer, as in Eqn.~\ref{eqn:lat-3}. Darker colors maps to greater accumulated attention scores. In this example, the input token ``bad'' receives the highest attention score.}
    \label{fig:lat}
\end{figure}

To study whether structured attention weights encode semantics, we propose a tracing method, Layer-wise Attention Tracing (LAT), to trace the attention `paid' to input tokens (i.e. words) through the self-attention layers in our encoder. LAT, illustrated in Fig.~\ref{fig:lat}, involves three main steps. First, starting from the local attention layer and a fixed ``quantity'' of attention, we distribute attention weights back to $\textbf{\textit{Z}}_{h}^{L}$, the last self-attention layer of each head $h \in [1, H]$. Second, we trace the attention back through each self-attention layer $l \in [1, L]$. Third, from the first layer of each head, we trace the attention back onto each token in the input sequence, by accumulating attention scores from each head to the corresponding position. 
We do not consider the decoder in LAT, as the MLP and LSTM layers in the decoder do not modify attention. Furthermore, we specifically ignore the feedforward layers and residual connections in the encoder, as we were interested in the attention $\boldsymbol\alpha$, not the neural activations they modify---this is our main differentiation from gradient-based or relevance-based work \cite{voita-etal-2019-analyzing}, and we note another recent paper  \cite{abnar-zuidema-2020-quantifying} that made the same assumptions.

\paragraph{Tracing Local Attention.} Given an input sequence $\textbf{\textit{X}}$ of length $N$ tokens, the forward pass of the model
(Eqn.~\ref{eqn:first}-\ref{eqn:ctx-c}) 
transforms $\textbf{\textit{X}}$ into the context vector $c$. We consider how a fixed quantity of attention, $\textbf{\textit{A}}^{c}$, gets divided back to the various heads. We refer to this quantity as the \textit{Attention Score} that is accumulated down through the layers. From Eqn.~\ref{eqn:ffn} and Eqn.~\ref{eqn:ctx-c}, we note that $c$ is a function of concatenated $\textbf{\textit{Z}}^{L}$ from the last self-attention layer, from each of the heads:
\begin{align}
    c &= \sum_{i = 1}^{N} \alpha_{i}^{c} f_{\psi}^{L}([z^{L}_{1(i)}, ..., z^{L}_{H(i)}]) \label{eqn:lat-ctx}
\end{align}
where $z^{L}_{h(i)}$ is the attended \emph{Value} vector from head $h \in [1, H]$ of the last layer $L$ at position $i \in [1, N]$. On the forward pass, the contribution of head $h$ at position $i$, $z^{L}_{h(i)}$, is weighted by $\alpha^{\textit{c}}_{i}$; Thus, on this first step of LAT, we divide the attention score $\textbf{\textit{A}}^{c}$ back to head $h$ at position $i$, using $\alpha^{\textit{c}}_{i}$:
\begin{align}
    \textbf{\textit{A}}_{h(i)}^{L+1} = \alpha_{i}^{c} \textbf{\textit{A}}^{c} \label{eqn:lat-4}
\end{align}
We use this notation to allude that this is the attention weights coming down from the ``$(L+1)$-th layer'', to follow the logic of the next step of LAT. Without loss of generality, we can set the initial attention score at the top, $\textbf{\textit{A}}^{c}$, to be 1, then all subsequent attention scores can be interpreted as a proportion of the initial attention score. 
%
Note that in our attention tracing, we are interested in accumulating the attention $\textbf{\textit{A}}_{h(i)}^{l}$ for each layer $l \in [1, L]$ at each position $i$, and so we focus on the attention weights (and not the hidden states that the attention multiplies, $\textbf{\textit{Z}}_h$ or $\textbf{\textit{V}}_h$), which remain unchanged through $f_\psi$.


\paragraph{Tracing Self Attention.} 
On the forward pass, Eqn. \ref{eqn:self-v} applies the self-attention weights. We rewrite this equation to make the indices explicit:
\begin{align}
    z_{h(j)}^{l} &= \sum_{i = 1}^{N} \alpha_{h(i\rightarrow  j)}^{s(l)} v_{h(i)}^{l} \label{eqn:lat-1}
\end{align}
where $v_{h(i)}^{l}$ denotes the $i$-th row of $\textbf{\textit{V}}_{h}^{l}$ (i.e., corresponding to the token in position $i$), and $\alpha_{h(i\rightarrow  j)}^{s(l)}$ is the $(j, i)$ element of $\boldsymbol\alpha^{\textit{s}(l)}_{h}$, such that it captures the attention from position $i$ to position $j$. 
The attended values $\textbf{\textit{Z}}_{h}^{l}$ then undergo two sets of feed-forward layers: Eqn.~\ref{eqn:ffn} with $f_\psi^l$ to get $\textbf{\textit{X}}^{l}$ and Eqn.~\ref{eqn:first} with $f_v^l$ to get $\textbf{\textit{V}}_{h}^{l+1}$. 

Using $\textbf{\textit{A}}_{h(j)}^{l}$ to denote the attention score accumulated at head $h$, position $j$, layer $l$, we can trace the attention coming down from the next-higher layer based on Eqn. \ref{eqn:lat-1}: 
%
\begin{align}
    \textbf{\textit{A}}_{h(i)}^{l} &= \sum_{j = 1}^{N} \alpha_{h(i\rightarrow  j)}^{s(l)} \textbf{\textit{A}}_{h(j)}^{l+1} \label{eqn:lat-3}
\end{align}
To confirm our intuition, on the forward pass (see Eqn. \ref{eqn:lat-1} and Fig. \ref{fig:lat}), to get the hidden value at position $j$ on the ``upper" part of the layer, we sum $\alpha_{h(i\rightarrow  j)}^{s(l)}$ over $i$ (the indices of the ``lower" layer). Thus, on the LAT pass downwards (Eqn. \ref{eqn:lat-3}), to get $\textbf{\textit{A}}_{h(i)}^{l}$ as position $i$ on the ``lower" layer, we sum the corresponding $\alpha_{h(i\rightarrow  j)}^{s(l)}$'s over $j$.


\noindent
\textbf{Tracing to input tokens. }Finally, for each input token $X_i$, we sum up the attention weights from each head at the corresponding position in the first layer to obtain the accumulated attention weights paid to token $X_i$:
\begin{align}
    \textbf{\textit{A}}_{i}^{x} &= \sum_{h = 1}^{H} \textbf{\textit{A}}_{h(i)}^{1} \label{eqn:lat-last} 
\end{align}
In summary, Eqns. \ref{eqn:lat-4}, \ref{eqn:lat-3}, and \ref{eqn:lat-last} describe the LAT method for tracing through the local and self-attention layers back to the input tokens $X_i$.

\section{Related Work}

There has been extensive debate over what attention mechanisms learn. On the one hand, researchers have developed methods to probe learnt self-attention in Transformer-based models, and show that attention scores learnt by models like BERT encode syntactic information like Parts-of-Speech~\cite{vig2019analyzing}, dependencies~\cite{hewitt2019structural, raganato2018analysis}, anaphora~\cite{goldberg2019assessing, voita2018context} and other parts of the traditional NLP pipeline~\cite{tenney2019bert}. These studies collectively suggest that self-attention mechanisms learn to encode syntactic information, which led us to propose the current work on whether self-attention can similarly learn to encode \emph{semantics}.

On the other hand, there are also other papers questioning the interpretations the field has placed on attention. These researchers show that attention weights have a low correlation with gradient-based measures of importance~\cite{jain2019attention, serrano2019attention, vashishth2019attention}. More recent analysis suggest that in certain regimes for the Transformer (i.e., sequence length greater than attention head dimension\footnote{We note that our models do not fall into this regime.}), attention distributions are non-identifiable, posing problems for interpretability~\cite{brunner2020identifiability}. 
In our work, we provide a method that can trace attention scores in Transformers to the input tokens, and show with both qualitative and quantitative evidence that these scores are semantically meaningful.

Beyond attention-based studies, there have been numerous studies that proposed gradient-based attribution analyses~\cite{dimopoulos1995use, gevrey2003review, Simonyan2014DeepIC} and layer-wise relevance propagation~\cite{bach2015pixel, li2016visualizing, arras2017explaining}. Most related to the current work is \citet{voita-etal-2019-analyzing}, who extended layer-wise relevance propagation to the Transformer to examine the contribution of individual heads to the final decision. In parallel, \citet{abnar-zuidema-2020-quantifying} recently proposed a method to roll-out structured attention weights inside the Transformer model, which is similar to our LAT method we propose here, although we provided more analysis via an external validation using external knowledge. We sought to investigate the attention accumulated onto individual input tokens using attention tracing, in a more similar manner to \citet{vig2019analyzing} for syntax or how \citet{voita2018context} looked at the attention paid to other words. We also calculate a gradient-based score (see Eqn. \ref{eqn:gradient-score}) to contrast our attention results with, and though these two scores are correlated (see Footnote \ref{footnote:gradient-attention-correlation}), they behave differently in our analyses. 

\section{Datasets}\label{sec:dataset}

\begin{table}[!t]
\centering
    {
    \setlength\tabcolsep{2.5pt}
    \begin{tabular}{p{4.2cm}p{1cm}}
    \toprule
    \multirow{2}{*}{\textbf{Model}} & \multicolumn{1}{c}{SST-5} \\ \cmidrule(l){2-2} 
     & \multicolumn{1}{c}{Accuracy (SD$^{\text{runs}}$)} \\ \midrule
    RNTN~\cite{socher2013recursive} & \multicolumn{1}{c}{45.7 (-)} \\
    BiLSTM~\cite{tai2015improved} & \multicolumn{1}{c}{46.5 (-)}   \\
    {\small Transformer + Position encoding} ~\cite{ambartsoumian2018self} & \multicolumn{1}{c}{\multirow{2}{*}{45.0 (0.4)}}\\
    DiSAN~\cite{shen2018disan} & \multicolumn{1}{c}{51.7 (-)}  \\ 
    \hline
    \textit{Our Self-attention} & \multicolumn{1}{c}{47.5 (0.2)}   \\ \midrule
    
   \multirow{2}{*}{\textbf{}} & \multicolumn{1}{c}{SEND} \\ \cmidrule(l){2-2}
   & \multicolumn{1}{c}{CCC (SD$^{\text{runs}}$, SD$^{\text{eg}}$)} \\ \midrule
    
    LSTM~\cite{ong2019modeling} & \multicolumn{1}{c}{.40 (-, .32)} \\
    SFT~\cite{wu2019attending} & \multicolumn{1}{c}{.34 (-, .33)} \\
    Human~\cite{ong2019modeling} &  \multicolumn{1}{c}{.50 (-, .12)}   \\ 
    \hline
    \textit{Our Self-attention + LSTM} & \multicolumn{1}{c}{.54 (.02, .36)}   \\ \midrule 
    \bottomrule
    \end{tabular}
    }
    \vspace{3pt}
    \caption{Summary of results. Top: Test accuracy averages and standard deviations (in brackets) for SST-5. Bottom: Test CCC averages and standard deviations on the SEND Test set. We additional calculate a SD$^\text{eg}$ over the CCCs of the same model over the (39) examples in the Test set, as used in previous papers, to better estimate generalizability to new unseen examples.} 
    \label{tab:SummaryOfResults}
\end{table}

To show that our interpretation methods can generalize across different types of datasets, we apply our method to two tasks with different characteristics, namely, sentiment classification of movie reviews on the Stanford Sentiment Treebank (SST), and time-series valence regression over long sequences narrative stories on the Stanford Emotional Narratives Dataset (SEND).

\subsection{Stanford Sentiment Treebank}


We used the fine-grained (5-class) version of the Stanford Sentiment Treebank (SST-5) movie review dataset ~\cite{socher2013recursive}, which has been used in previous studies of interpretability of neural network models ~\cite{li2016visualizing, arras2017explaining}. All sentences\footnote{Although the SST contains labels on each parse tree of the reviews, we only considered full sentences.} were tokenized, and preprocessed by lowercasing, similar to ~\cite{li2016visualizing}. We embed each token using 300-dimensional GloVe word embeddings \cite{pennington2014glove}. 
Each sentence is labeled via crowdsourcing with one of five sentiment classes \{Very Negative, Negative, Neutral, Positive, and Very Positive\}. We used the same dataset partitions as in the original paper: a \textit{Train set} (8544 sentences, average length 19 tokens), a \textit{Validation set} (1101 sentences, average length 19 tokens) and a \textit{Test set} (2210 sentences, average length 19 tokens). Models are trained to maximize the 5-class classification accuracy by minimizing multi-class cross-entropy loss. We compare our model with previous works on SST that are based on LSTM~\cite{tai2015improved} and Transformer~\cite{ambartsoumian2018self, shen2018disan}.



\subsection{Stanford Emotional Narratives Dataset}

The SEND~\cite{ong2019modeling} comprises videos of participants narrating emotional life events. Each video is professionally transcribed, and annotated via crowdsourcing with emotion valence scores ranging from ``Very Negative'' [-1] to ``Very Positive'' [1] continuously sampled at every 0.5s. Details can be found on the authors' GitHub repository.
The SEND has previously been used to train deep learning models to predict emotion valence over time~\cite{ong2019modeling, wu2019attending}. 

The SEND has 193 transcripts, and each one contains multiple sentences. We preprocess them by tokenizing and lowercasing as in~\cite{ong2019modeling, wu2019attending}. Additionally, we divide each transcript into 5-second time windows by using timestamps provided in the dataset. We use the average valence scores during a time window as the label of that window. We use the same partitions as in the original paper: a \textit{Train set} (114 transcripts, average length 357 tokens, average window length 13 tokens), a \textit{Validation set} (40 transcripts, average length 387 tokens, average window length 15 tokens) and a \textit{Test set} (39 transcripts, average length 333 tokens, average window length 13 tokens). We embed each token in the same way as for SST-5. As in the original papers~\cite{ong2019modeling, wu2019attending}, we use the Concordance Correlation Coefficient (CCC \cite{lin1989concordance}) as our evaluation metric (See Appendix for the definiton). We compare our model with previous works on SEND that use LSTM~\cite{ong2019modeling} and Transformer~\cite{wu2019attending}.

\section{Results and Analysis}\label{sec:analysis}

\subsection{Model training and results}

We report the results of our Transformer-based models in Table~\ref{tab:SummaryOfResults} with performances of state-of-the-art (SOTA) models trained with these two datasets. We selected models in the literature that are the most representative and relevant to our models. Our Transformer-based model for the SST-5 classification task (Fig.~\ref{fig:model}) achieves good performance, with an accuracy ($\pm$ standard deviation) of 47.5\% $ \pm $ 49.9\% on the five-class sentiment classification. For the SEND dataset, our model outperforms previous SOTA models and even average human performance on this task, with a mean CCC of .54 $\pm$ .36 on the Test set. Interestingly, our window-based Transformer encoder increases performance compared to the Simple Fusion Transformer proposed by~\citet{wu2019attending}, who used a Transformer-based encoder over the whole narrative sequence.

\begin{figure*}[h]
    \centering
    \includegraphics[width=0.96\textwidth]{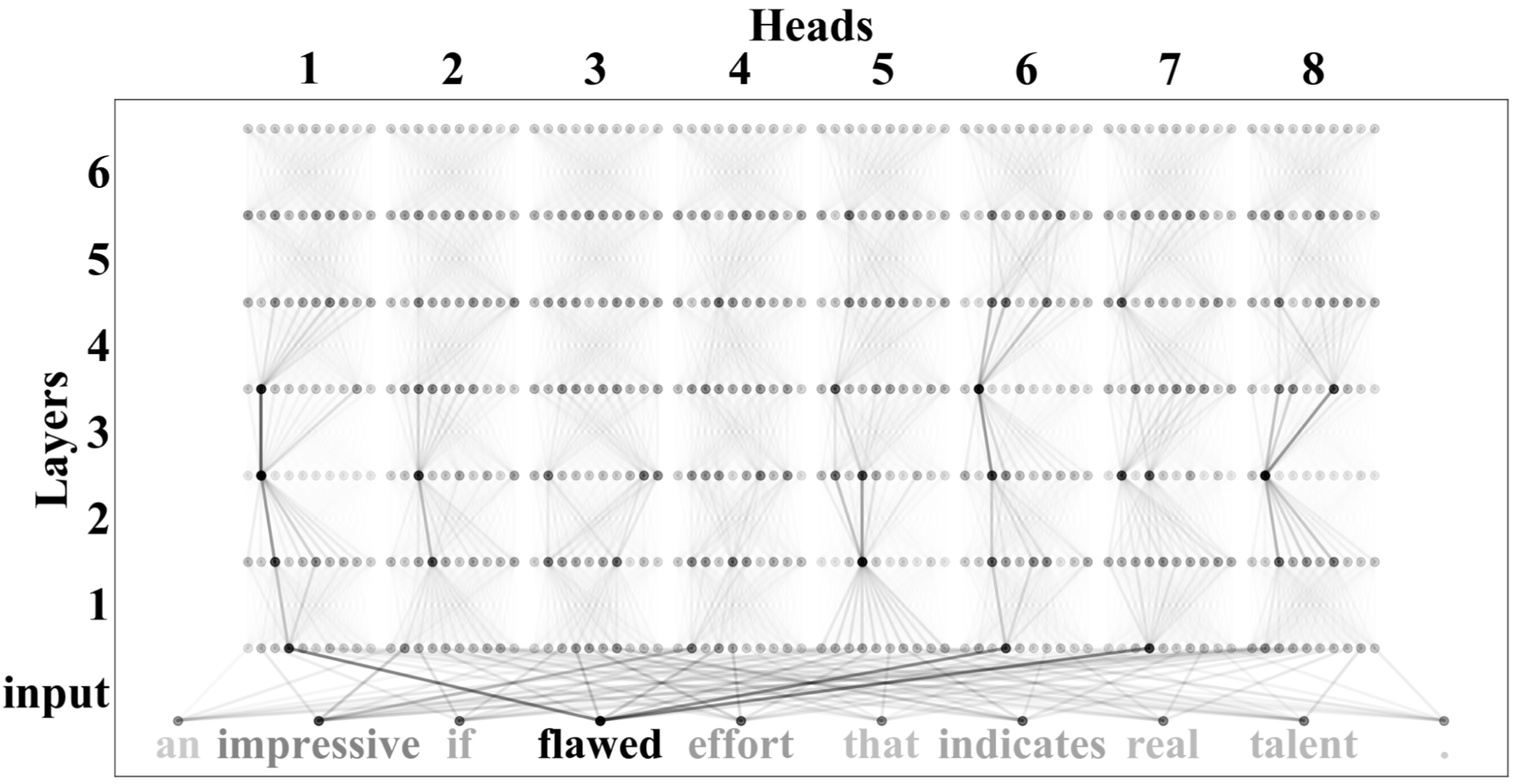}
    \caption{An example \emph{flow diagram} of attention distributed through self-attention layers in action. On the bottom, the font weights illustrate the accumulated attention weights paid to a particular word. The predicted label and the true label are both \emph{positive}. Note that the color of the dots represent the attention weights $\textbf{\textit{A}}_{h(i)}^{l}$ (Eqn. 10), not the activation of those neurons, and so these are not affected by the states that are shared across heads.
    }
    \label{fig:flow}
\end{figure*} 

Both models are trained with the \emph{Adam}~\cite{DBLP:journals/corr/KingmaB14} optimization algorithm with a learning rate of $10^{-4}$. As our goal was analyzing structured attention weights, not maximizing performance, we manually specified hyperparameters without any grid search. We include details about our experiment setup in the Appendix.

Given that our Transformer-based models achieved comparable state-of-the-art performance on the SST and SEND, we then proceed to analyze the attention scores produced by LAT on these models. After computing $\textbf{\textit{A}}_{i}^{x}$ for all the words in a given sequence, we normalize attention scores using the softmax function to have them sum to 1.



\subsection{Visualizing Layerwise Attention Tracing}

The \emph{flow diagram} in Fig.~\ref{fig:flow} visualizes how attention aggregates using LAT across all heads and layers for the model trained with SST-5 for an example input. Rows represents self-attention layers and columns represent attention heads. Dots represent different tokens at head $h \in [1, H]$ (left to right), position $i \in [1, N]$ of layer $l \in [1, L]$ (bottom to top). Dots in the bottom-most layer represents input tokens.
The darker the color of each dot, the higher the accumulated attention score at that position, calculated using by Eqns.~\ref{eqn:lat-4}, ~\ref{eqn:lat-3} and ~\ref{eqn:lat-last}. Attention weights $\alpha_{h(i\rightarrow  j)}^{s(l)}$ in each layer are illustrated by lines connecting tokens in consecutive layers.

This diagram illustrates some coarse-grained differences between heads. For example, all heads in the top last layer distributed attention fairly equally across all tokens.
Other heads (e.g., Head 6, Layer 4, and Head 8, Layer 3) have a downward-triangle pattern, where attention weights are accumulated to a specific token in a lower-layer, while others (e.g. Head 5, Layer 1) seem to re-distribute accumulated attention more broadly.
Finally, at the input layer, we note that attention scores seem to be highest for words with strong emotion semantics.

\subsection{Sentiment Representations of Words}
\begin{figure}
\centering
\begin{subfigure}{.24\textwidth}
  \centering
  \includegraphics[width=1\linewidth]{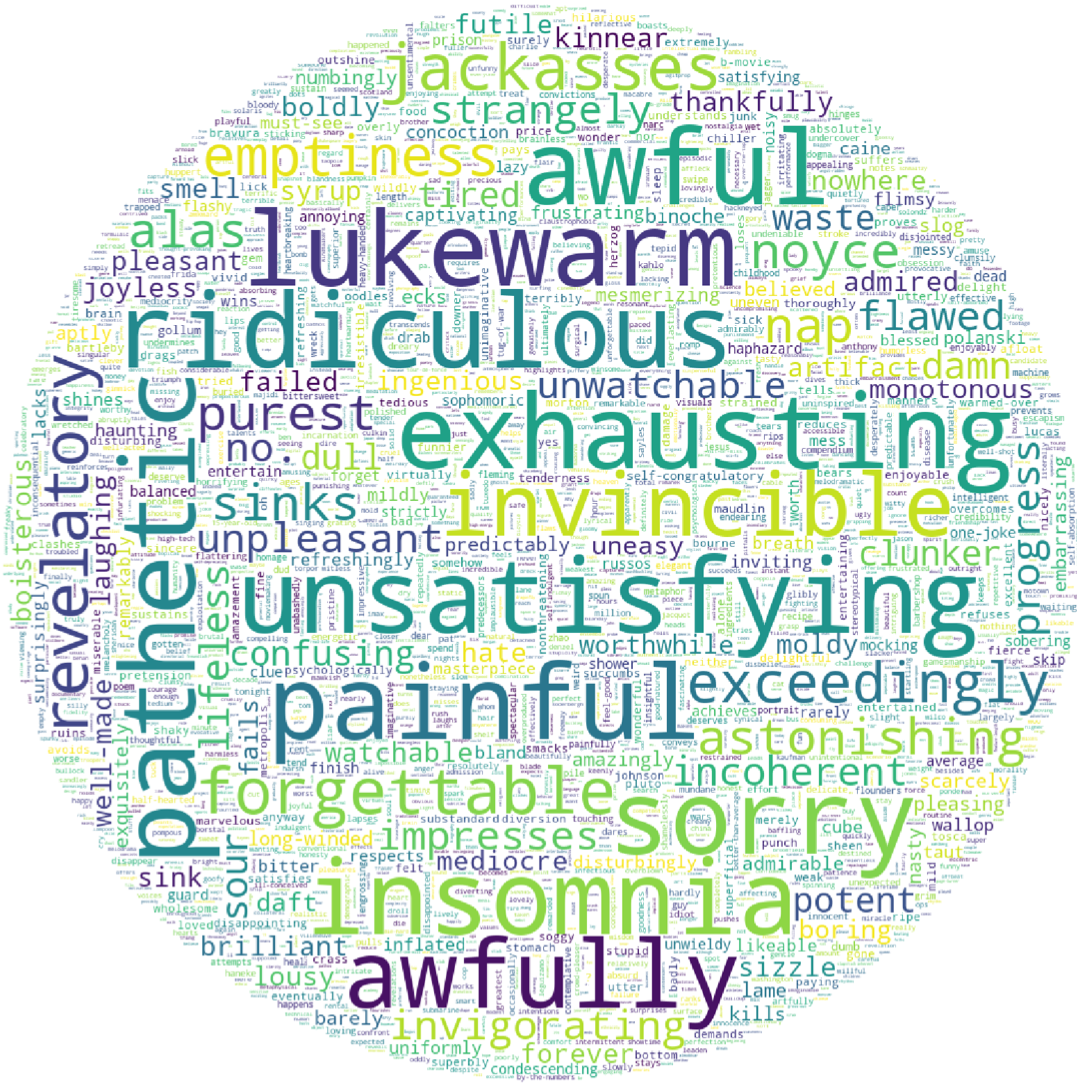}
  \caption{SST-5.}
  \label{fig:sub1}
\end{subfigure}%
\begin{subfigure}{.24\textwidth}
  \centering
  \includegraphics[width=1\linewidth]{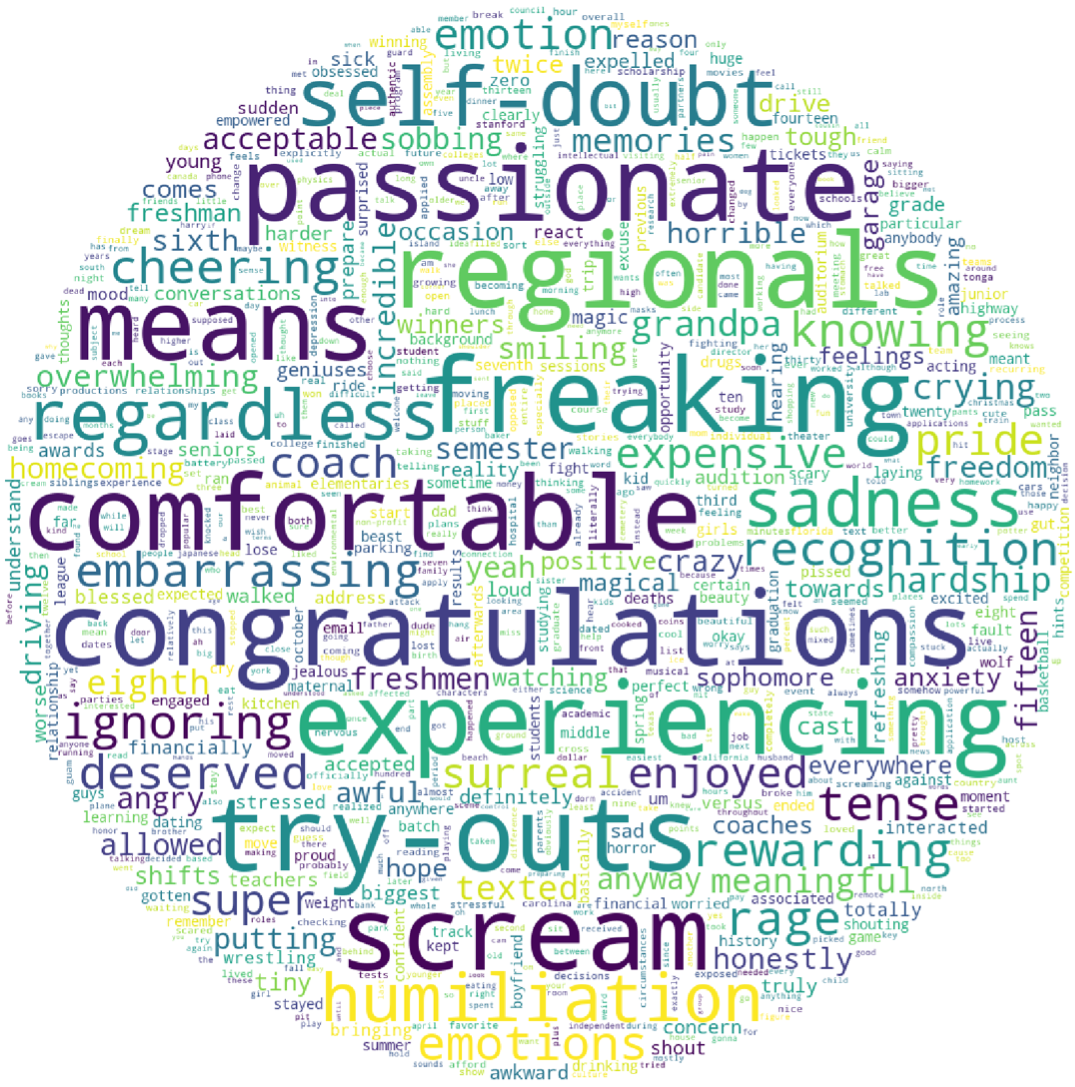}
  \caption{SEND.}
  \label{fig:sub2}
\end{subfigure}
\caption{Word cloud created based on averaged accumulated attention weights assigned to words in the vocabularies of Test sets.}
\label{fig:test}
\end{figure}

To validate that the attention weights aggregated on the input tokens by LAT is semantically meaningful, we rank all unique word-level tokens in the Test set by their averaged attention scores received from all sequences that they appear. Concretely, we first use LAT to trace attention weights paid to input tokens for every sequences in the Test set. For tokens that appear more than once, we average their attention scores across occurrences. We then rank tokens by their average attention score, and illustrate in Fig.~\ref{fig:test} using word clouds where a larger font size maps to a higher average attention score. For both datasets, we observe that words expressing strong emotions also have higher attention scores, see e.g. \emph{sorry}, \emph{painful}, \emph{unsatisfying} for SST-5, and  \emph{congratulations}, \emph{freaking}, \emph{comfortable} for SEND. We note that stop words do not receive high attention scores in either of the datasets.

\subsection{Quantitative validation with an emotion lexicon}

\begin{figure}
\centering

\includegraphics[width=1\columnwidth]{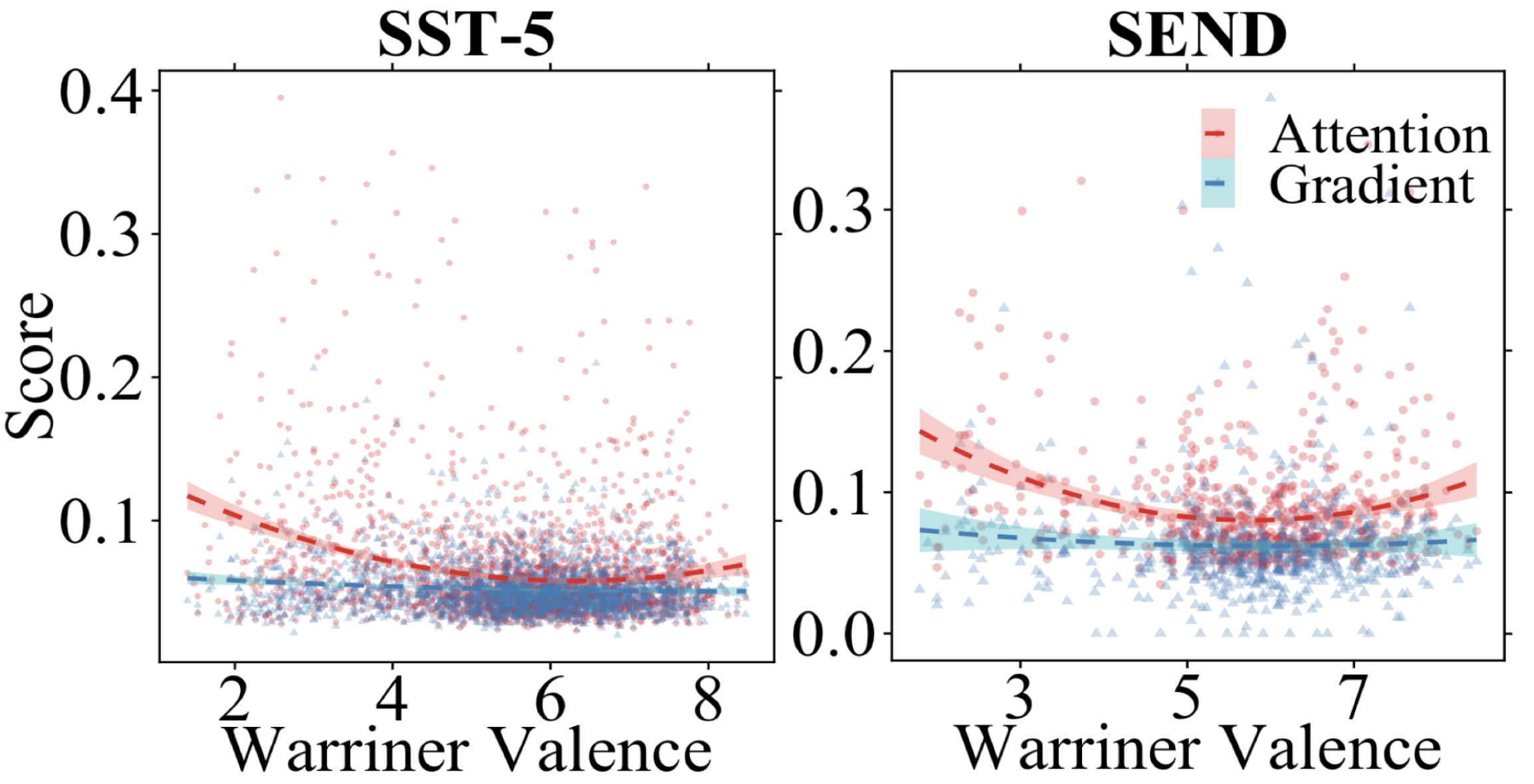}
\caption{Scatterplot shows scores on y-axis derived from Eqn.~\ref{eqn:war} (LAT attention scores $\textbf{\textit{A}}_{w}$ in red circles and gradient scores $\textbf{\textit{G}}_{w}$ in blue triangles) and corresponding emotional valence ratings, $\text{Val}_{w}$, from the \citet{warriner2013norms} lexicon on x-axis. Shared vocabulary size is 2335 for SST-5, and 660 for SEND.}
  \label{fig:u-shape}
\end{figure}

One advantage of extracting emotion semantics from natural language text is that the field has amassed large, annotated references of emotion semantics. We refer, of course, to the emotion lexicons that earlier NLP researchers used for sentiment analysis and related tasks \cite{hu2004mining}. Although they seem to have fallen out of favor with the rise of deep learning (and the hypothesis that deep learning can learn such knowledge in a data-driven manner), in our task, we sought to use emotion lexicons as an external validation of what our model learns.

We used a lexicon \cite{warriner2013norms} of nearly 14,000 English lemmas that are each annotated by an average of 20 volunteers for emotional valence, which corresponds exactly to the semantics in our tasks. The mean valence ratings in this lexicon are real-valued numbers from 1 to 9.

We hypothesize that our LAT method produce attention scores such that words having higher scores will tend to have greater emotional meaning. Additionally, since our attention scores do not differentiate emotion ``directions'' (i.e., negative and positive), these attention scores should be high for both very positive words, \emph{as well as} very negative words. Thus, we expect a \textit{U-shaped} relationship between our attention scores and the lexicon's valence ratings. 
We examine this hypothesis by fitting a quadratic regression equation\footnote{Specifically, we used the following formula in R syntax: \texttt{lm(att $\sim$ poly(val,2))}, where \texttt{poly()} creates orthogonal polynomials to avoid collinearity issues.}:
\begin{align}
    \textbf{\textit{A}}_{w} = b_0 + b_1 \text{Val}_{w} + b_2 [\text{Val}_{w}]^2 + \epsilon \label{eqn:war}
\end{align}
where $\textbf{\textit{A}}_{w}$ is the averaged attention score of a particular word $w$ derived by the LAT method, and $\text{Val}_{w}$ represents the valence rating of that word from the \citet{warriner2013norms} lexicon. We hypothesized a statistically-significant coefficient $b_2$ on the quadratic term. 

To contrast our attention score with another measure of importance, the gradient, i.e., how important the inputs are to affecting the output~\cite{li2016visualizing}, we also calculate a gradient score on each token by computing squared partial derivatives:
\begin{align}
    \textbf{\textit{G}}_{w(d)} &= (\pd{f_{\xi}}{w_{(d)}} (w))^{2} \label{eqn:gradient-score}
\end{align}
where $f_{\xi}$ can be parameterized by neural networks, and $\textbf{\textit{G}}_{w(d)}$ is the gradient of a particular space dimension $d$ of the embedding for the input token $w$. We then regress $\textbf{\textit{G}}_{w}$ on the lexicon valence ratings using Eqn.~\ref{eqn:war}. 



We plot both our attention scores and gradient scores for each word against \citet{warriner2013norms} valence ratings, in Fig.~\ref{fig:u-shape}. For both tasks, we considered only words that appeared in both our Test sets and the lexicon, and plot only scores below 0.4 to make the plot more readable\footnote{This plotting rule only filtered out less than 1\% of words in the Test sets: .171\% for SST-5 and .754\% for SEND.}. We can see clearly that there exist a U-shaped, quadratic relationship between attention scores and the Warriner valence ratings ($b_2 =  0.283, SE=0.040, t=7.04, p < .001$ for SST-5; $b_2 = 0.242, SE=0.039, t=6.21, p < .001$ for SEND). Our results support our hypothesis that the attention scores recovered by our LAT method do track emotional semantics. As a result, we show that structured attention weights may encode semantics independent of other types of connections in the model (e.g., linear feedforward layers and residual layers.).
%
By contrast, there is no clear quadratic relationship between gradient scores and valence ratings across both tasks (SST-5, $p=0.19$; SEND, $p=0.28$)\footnote{On the SST, $\textbf{\textit{A}}_{i}^{x}$ and $\textbf{\textit{G}}_{i(d)}^{x}$ are correlated at $\rho=.80$, and on the SEND, $\rho=.37$. The two values are highly correlated (on the SST), but vary differently with respect to valence. \label{footnote:gradient-attention-correlation}}. 


\subsection{Head Attention on Sentiment Words}
\begin{figure}
    \centering
    \includegraphics[width=0.48\textwidth]{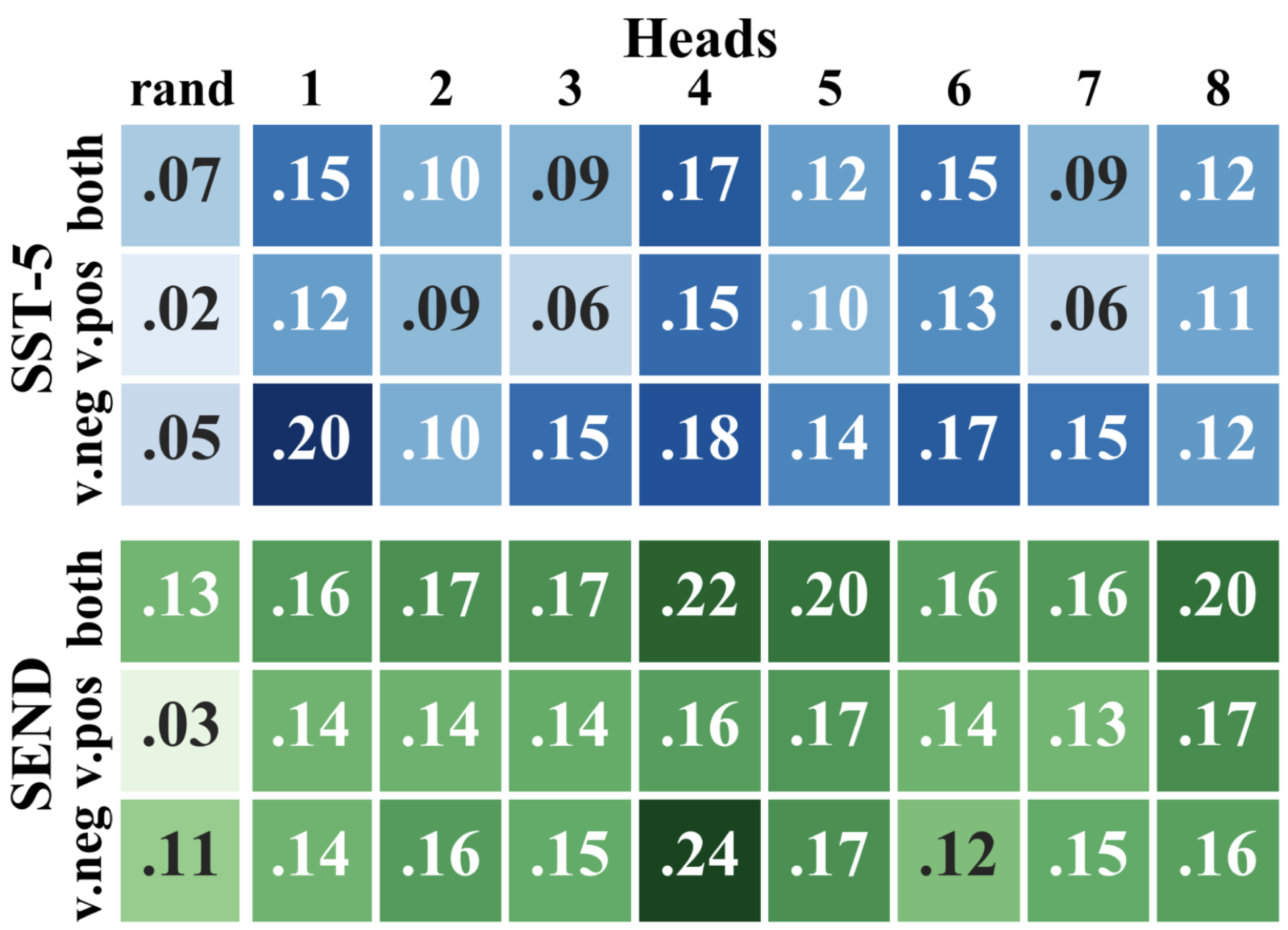}
    \caption{Heatmap of proportions of attention paid to words with selected semantics tags. The leftmost column ``rand'' shows the proportions if attention weights are uniformly distributed at chance.} 
    \label{fig:heatmap}
\end{figure}

We next analyze the amount of attention paid to sentiment words in each head. Within each head $h$, we analyze the proportion of accumulated attention $\textbf{\textit{A}}_{h(i)}^{1}$ on emotional words, specifically focusing on \emph{very positive} and \emph{very negative} words\footnote{For SST-5, we used the original word-level \emph{very positive} and \emph{very negative} labels in the dataset. For SEND, we used the Warriner lexicon and chose a cutoff $\geq 6.5$ for very positive, and $< 3.5$ for very negative.}, aggregated over the Test sets:
%
%
\begin{align}
    p_{\textbf{\textit{A}}_{h}^{1}}(\text{tag}) = \frac{ \mathlarger{\sum}\limits_{\textbf{\textit{X}} \in \mathcal{X} } \mathlarger{\sum}_{i=1}^{|\textbf{\textit{X}}|} (\textbf{\textit{A}}_{h(i)}^{1} ) \mathds{1}_{\text{label}(x_{i}) = \text{tag}} }{\mathlarger{\sum}\limits_{\textbf{\textit{X}} \in \mathcal{X} } \mathlarger{\sum}_{i=1}^{|\textbf{\textit{X}}|} (\textbf{\textit{A}}_{h(i)}^{1} ) }
\end{align}
where $\mathcal{X}$ is the subset of sequences that contain at least 1 word with the selected tag\footnote{That is, when calculating $p_{\textbf{\textit{A}}_{h}^{1}}(\emph{very positive})$, we exclude sequences that do not contain at least 1 \emph{very positive} word.}.

Fig.~\ref{fig:heatmap} shows the proportion of attention accumulated by heads to \emph{very positive} and \emph{very negative} words, compared with chance. 
All heads do seem to pay more attention to strongly emotional words, compared to chance, and some heads seem to `specialize' more: For example, Head 4 in our SEND model pays 24\% of its accumulated attention to \emph{very negative} words while the mean of all other heads is closer to 15\%. 
While Fig.~\ref{fig:heatmap} is specific to the model we trained, it is illustrative that specialization to strong emotional semantics does emerge from the learnt attention weights.

\section{Discussion}\label{sec:discussion}
In this work, we analyzed whether structured attention weights encode semantics in sentiment analysis tasks, using our proposed probing method LAT to trace attention through multiple layers in the Transformer. We demonstrated that the accumulated attention scores tended to favor words with greater semantic meaning, in this case, emotional meaning. We applied LAT to two tasks having similar semantics, and show that our results generalize across both tasks/domains. We validated our results quantitatively with an emotion lexicon, and showed that our attention scores are highest for both highly positive and highly negative words---our \emph{a priori} hypothesis for the quadratic, ``U-shaped'' relationship. We also found some evidence for specialization of heads to emotional meaning. Although it may seem that our attention tracing is ``incomplete'' as it does not take into account the feed-forward layers and residual connections, by contrast, this quadratic relationship was not shown by pure gradient-based importance, which suggests that there may be some utility to looking only at attention.

We believe that attention in its various forms \cite{luong2015effective, vaswani2017attention} are not only effective for performance, but may also provide interpretable explanations of model behaviour. It may not happen with today's implementations; we may need to engineer inductive biases to constrain attention mechanisms in order to address issues of identifiability that \citet{jain2019attention} and others have pointed out. And perhaps, attention should not be interpreted like gradient-based measures (see Fig. \ref{fig:u-shape}). This debate is not yet resolved, and we hope our contributions will be useful in informing future work on this topic.


\bibliography{emnlp2020}
\bibliographystyle{acl_natbib}

\newpage

\section{Supplementary Materials}

\subsection{Evaluation Metrics}

\paragraph{Concordance Correlation Coefficient (CCC \cite{lin1989concordance}): } The CCC of vectors $X$ and $Y$ is:
\begin{align}
    \text{CCC}(X,Y) &\equiv 
    \frac{2 \text{Corr}\left(X,Y\right) \sigma_X \sigma_Y}{\sigma_X^2 + \sigma_Y^2 + \left( \mu_X - \mu_Y \right)^2}
\end{align}
where $\text{Corr}\left(X,Y\right) \equiv \text{cov}(X,Y)/(\sigma_X \sigma_Y)$ is the Pearson correlation coefficient, and $\mu$ and $\sigma$ denotes the mean and standard deviation respectively.

\subsection{Experiment Setup}

\paragraph{Computing Infrastructure: } To train our models, we use a single Standard NV6 instance on Microsoft Azure. The instance is equipped with a single NVIDIA Tesla M60 GPU.

\paragraph{Average Runtime: } With the computing infrastructure, it takes about 1.5 hrs to train both models, where each model is trained with 200 epochs. Both model reaches maximum performances in about 1.5 hrs at about 100 epochs.

\paragraph{Number of Trainable Parameters: } The model trained on SST-5 that uses a LSTM decoder has 3,993,222 parameters. Additionally, the model trained on SEND that uses a MLP decoder has 4,715,362 parameters.

\subsection{Task-specific Decoders}
\label{sec:supplemental}

\paragraph{Long-Short Term Memory Network (LSTM): } For the time-series task, we use a LSTM layer~\cite{hochreiter1997long} to decode the context vector $c_{i}$ from our encoder for each window $i$ to output a hidden vector $h_{i}$. Then, the hidden vector passes through a MLP to make the valence prediction:
\begin{align}
    h_{i} &= \text{LSTM}(h_{i-1}, c_{i}) \label{eqn:LSTM} \\
    \hat{r}_{i} &= \text{MLP}(h_{i})
\end{align}

\paragraph{Multilayer Perceptron (MLP): } Our MLP contains 3 consecutive linear layers with a single ReLU activation in between layers. For the classification task, we feed in the context vector $c$ from our encoder to MLP to make the sentiment prediction:
\begin{align}
    f_1(c) &= \text{ReLU}(\textbf{W}_{1} c + \textbf{b}_{1}) \\
    \hat{r}_{i} &= \textbf{W}_3 \left[ \text{ReLU}(\textbf{W}_{2} f_1(c) + \textbf{b}_{2}) \right] + \textbf{b}_3
\end{align}
where $\textbf{W}_{1}, \textbf{W}_{2}, \textbf{W}_{3}, \textbf{b}_{1}, \textbf{b}_{2}, \textbf{b}_{3}$ are learnable parameters for linear layers. For the time-series task, the hidden vector $h_t$ is the input instead.

\end{document}